\documentclass[runningheads]{llncs}
\usepackage[a4paper, margin=1in]{geometry}
\usepackage{array} 
\usepackage[table]{xcolor}
\usepackage{cite}
\usepackage[T1]{fontenc}    
\usepackage{graphicx}
\usepackage{color}
\usepackage{hyperref}
\usepackage[utf8]{inputenc}  
\usepackage{amsmath}
\usepackage{mathtools}
\usepackage{amsfonts}
\usepackage{multirow}
\usepackage{adjustbox}
\usepackage{booktabs}
\usepackage{enumerate} 
\usepackage{hyperref}       
\usepackage{amsfonts}       
\usepackage{parskip}
\usepackage{nicefrac}       
\usepackage{float}

\begin{document}

\title{Empowering Medical Equipment Sustainability in Low-Resource Settings: An AI-Powered Diagnostic and Support Platform for Biomedical Technicians
}

\titlerunning{INGENZI Tech}

\author{Bernes Lorier Atabonfack\inst{1}\and
Ahmed Tahiru Issah\inst{1} \and
Mohammed Hardi Abdul Baaki\inst{2} \and Clemence INGABIRE \inst{2}\and
Tolulope Olusuyi\inst{3} \and Maruf Adewole, Dr\inst{4}  \and Udunna Anazodo, Dr\inst{4}\and Timothy Brown \inst{4}} 
\authorrunning{Bernes et al.}


\institute{Carnegie Mellon University Africa, Kigali, Rwanda \and
Medical Artificial Intelligence Laboratory, Lagos, Nigeria \and 
University of Pennsylvania \and
McGill University, Montr{\'e}al, Canada\\
\email{batabonf@andrew.cmu.edu},
\email{aissah@andrew.cmu.edu},
\email{mabdulba@andrew.cmu.edu},
\email{cingabir@andrew.cmu.edu},
\email{thetolusuyi@gmail.com},
\email{marufadewole@outlook.com},
\email{udunna.anazodo@mcgill.ca}, 
\email{timxb@cmu.edu}}

\maketitle              

\begin{abstract}
In low- and middle-income countries (LMICs), a significant proportion of medical diagnostic equipment remains underutilized or non-functional due to a lack of timely maintenance, limited access to technical expertise, and minimal support from manufacturers, particularly for devices acquired through third-party vendors or donations. This challenge contributes to increased equipment downtime, delayed diagnoses, and compromised patient care. This research explores the development and validation of an AI-powered support platform designed to assist biomedical technicians in diagnosing and repairing medical devices in real-time. The system integrates a large language model (LLM) with a user-friendly web interface, enabling imaging technologists/radiographers and biomedical technicians to input error codes or device symptoms and receive accurate, step-by-step troubleshooting guidance. The platform also includes a global peer-to-peer discussion forum to support knowledge exchange and provide additional context for rare or undocumented issues. A proof of concept was developed using the Philips HDI 5000 ultrasound machine, achieving 100\% precision in error code interpretation and 80\% accuracy in suggesting corrective actions. This study demonstrates the feasibility and potential of AI-driven systems to support medical device maintenance, with the aim of reducing equipment downtime to improve healthcare delivery in resource-constrained environments.
\end{abstract}

\section{Introduction and Background}

Medical devices are indispensable for delivering quality healthcare services, encompassing diagnosis, treatment, and patient monitoring. In high-resource settings, these devices benefit from regular maintenance, trained operators, and manufacturer support. Conversely, in low- and middle-income countries (LMICs), the situation is markedly different.
Studies estimate that 40\% to 70\% of medical equipment in LMICs is non-functional or underutilized at any given time due to factors like poor maintenance systems, lack of spare parts, and insufficiently trained personnel \cite{DiaconuKarin2017Mfmd} \cite{39663856} \cite{MalkinRobertA.2007DoHC} \cite{AnazodoUdunnaC.2023Affa}. For instance, a detailed case study in Tanzania found that 30–50\% of medical equipment in sub-Saharan Africa experiences downtime because of unstructured maintenance practices and limited technician capacity\cite{KebbyAbdallahAlly2024Mdmr}. In Uganda, research conducted across nine tertiary hospitals and five research institutions revealed that 34\% of medical devices were faulty, and 85.6\% lacked operational manuals, making even minor issues difficult to resolve \cite{SsekitolekoRobertTamale2022SoMD}. The lifespan of usable devices is often reduced by as much as 80\%, primarily due to inexperienced operators and a lack of preventive maintenance protocols \cite{MarksIsobelH2019Medi}.
The World Health Organization estimates that up to 70\% of donated equipment fails to function as intended due to mismatches with infrastructure or lack of user training \cite{10665-44568}. Moreover, 40\% of devices donated by Medical Surplus Recovery Organizations (MSROs) are inoperable due to missing documentation or limited manufacturer support\cite{Spring_2024} \cite{DiaconuKarin2014Mdpi}.

\par\hspace{2em} The shortage of trained biomedical engineering technicians (BMETs) further aggravates the issue. With only a few organizations globally offering formal BMET training (e.g., Engineering World Health, Medisend) \cite{Spring_2024}, many LMICs lack the workforce needed to keep devices operational. Additionally, equipment donated or procured through third-party vendors often comes without access to manufacturer service forums or troubleshooting support, leaving technicians isolated and ill-equipped.
In many African hospitals and clinics, even minor technical issues such as unfamiliar error codes or calibration warnings can result in devices being taken offline for extended periods. This leads to significant delays in diagnostics, unnecessary patient referrals, and, in some cases, avoidable mortality.
While digital health innovations have improved access to care and information, very few have directly addressed the operational gap in maintaining physical medical infrastructure. This study aims to bridge that gap through the design and implementation of an AI-powered medical device support system (INGENZI Tech) for diagnostic imaging devices. By integrating advanced natural language processing with structured manuals, log data, and peer-generated insights, the system empowers biomedical technicians to act with confidence. The goal of this project is to reduce downtime, improve technician efficiency, and enhance healthcare system reliability, starting with a validated use case on the Philips HDI 5000 ultrasound machine in East Africa.

\section{Related Work and Differentiation}

While a number of AI-driven platforms exist globally for medical device management, most are tied to proprietary hardware, operate in online-only contexts, or target high-resource health systems. Table 1 provides a comparative overview of notable commercial solutions and positions INGENZI Tech within this landscape. 

In parallel, academic research has made significant progress in predictive maintenance for medical equipment. For instance, Shamayleh et al.~\cite{ShamaylehAbdulrahim2020IBPM} deployed an IoT-based maintenance system across over 8,000 medical devices in Malaysia, achieving a 25\% cost reduction using SVM-based failure classification. Similarly, Zamzam et al.~\cite{ZamzamAizatHilmi2021ASRo} achieved 99.4\% prediction accuracy using sensor fusion methods applied to over 13,000 medical assets. Mohamed et al.~\cite{AbdWahabNurHaninie2024Srop} proposed a digital twin approach for MRI equipment, reducing machine downtime by 20\%. Deep learning approaches, such as CNN-LSTM hybrid models, have shown 92–99\% accuracy in remaining useful life estimation across several medical device categories~\cite{FernandesSofia2020FAFA, CinarZekiMurat2020MLiP}. Luschi et al.~\cite{GuissiMaroua2024IFPM} further demonstrated how predictive analytics integrated with IoT positioning systems can support hospital maintenance workflows in structured environments.

Despite this progress, such systems typically require continuous data streams, sensor infrastructure, and robust connectivity, conditions that are often absent in LMIC settings. Research by Anazodo et al.~\cite{AnazodoUdunnaC.2023Affa} and Diaconu et al.~\cite{DiaconuKarin2017Mfmd} highlights the infrastructural barriers to medical equipment usage in LMICs, including poor internet access, inconsistent power supply, and lack of OEM support, particularly for donated devices. 

In terms of knowledge management, Abidi~\cite{Abidi2007} and Tabrizi and Morgan~\cite{TabriziNegarMonazam2014MfDK} emphasize the need for collaborative technical knowledge platforms in healthcare. While prior work has demonstrated the effectiveness of training platforms~\cite{ArnesonWendy2013Blse}, few systems combine this with real-time, context-aware AI guidance for medical equipment.

\textbf{INGENZI Tech} differentiates itself through its integrated design tailored for LMICs. Unlike systems that focus solely on predictive analytics~\cite{ZamzamAizatHilmi2021ASRo, AbdWahabNurHaninie2024Srop} or manufacturer-side service 
workflows~\cite{a2025_circuitryaidecisionintelligencecompanyoverview}, INGENZI Tech offers an end-user diagnostic platform combining:
\begin{itemize}
    \item A multilingual, LLM-powered chatbot for step-by-step device guidance.
    \item A RAG-based architecture with segmented vector stores to reduce hallucinations and enhance contextual precision~\cite{LewisPatrick2020RGfK, FrielRobert2024REBf}.
    \item Offline-first deployment suitable for low-bandwidth and rural clinics.
    \item A peer-to-peer technician forum for continuous feedback, local adaptation, and model improvement~\cite{Abidi2007, TabriziNegarMonazam2014MfDK}.
\end{itemize}

The Phase 0 proof-of-concept targets the Philips HDI 5000 ultrasound, a device commonly found in African clinics but frequently unsupported, addressing the challenges raised in recent surveys on equipment non-functionality~\cite{SsekitolekoRobertTamale2022SoMD, 39663856}. Future phases include scaling to CT, MRI, and X-ray machines.

\begin{table}[htbp]
\centering
\caption{Comparative Landscape of Existing AI-Powered Medical Maintenance Solutions.}
\renewcommand{\arraystretch}{1.3}
\rowcolors{2}{gray!10}{white}
\setlength{\tabcolsep}{6pt}
\begin{tabular}{|p{3.0cm}|p{5.0cm}|p{6.5cm}|}
\hline
\rowcolor{gray!40}
\textbf{AI Solution} & \textbf{Strengths} & \textbf{Limitations in LMIC Context / Differentiation vs. INGENZI Tech} \\
\hline
\textbf{Bruviti} & LLM-powered diagnostics; preserves institutional knowledge; supports global manufacturers & No public deployments in LMICs; not optimized for offline use or multilingual environments; high dependency on OEM integrations \\
\hline
\textbf{Hadleigh Health, Vestfrost EMS, Nexleaf Analytics} & IoT-based predictive maintenance; real-time monitoring; some LMIC deployments & Primarily hardware-embedded systems; limited end-user diagnostic interaction; lacks LLM or multilingual troubleshooting support \\
\hline
\textbf{Bosnia AI-Metrology Project} & Applies AI to device calibration and performance in public hospitals; shows feasibility in low-resource settings & Regional, narrow scope; focused on measurement and standards, not interactive technician support or documentation retrieval \\
\hline
\textbf{Circuitry.ai} & Real-time advisory systems for maintenance; global operations; multilingual support & Tailored for manufacturers and enterprise clients; unclear adaptation to LMIC infrastructure or offline-first needs \\
\hline
\textbf{Stellarix, Kodexo Labs, Toronto Digital} & Offer AI-powered predictive maintenance with IoT and ML; adaptable for various device types & Require continuous data streams and sensor integrations; do not provide peer-to-peer learning or AI chatbot interactions \\
\hline
\textbf{GE Healthcare Edison AI} & Advanced anomaly detection in imaging systems; global footprint & High-end solution tied to GE devices; closed ecosystem; limited accessibility or relevance for third-party equipment in LMICs \\
\hline
\textbf{INGENZI Tech (This Work)} & LLM-based multilingual chatbot; RAG-based contextual retrieval; peer-driven technician forum; designed for offline use & Currently piloting; early-stage for multi-device support; open architecture enables adaptation across vendors and regions \\
\hline
\end{tabular}
\end{table}

\section{Scope of the Study}

This study focuses on the development and validation of an AI-powered support platform designed to assist biomedical technicians in diagnosing and repairing medical diagnostic equipment in low-resource settings. The platform integrates a large language model (LLM) with structured device knowledge, error logs, and technician-driven feedback to offer step-by-step troubleshooting and guided repair workflows. The initial scope centers on the Philips HDI 5000 ultrasound machine, which served as a proof-of-concept device for prototyping. With the framework developed in the initial phase, a full study will be conducted to expand the focus to include more complex imaging devices such as magnetic resonance imaging (MRI), computed tomography (CT), and X-ray machines, with a specific emphasis on widely used and often unsupported devices in LMIC settings. Therefore, this study aims to:
\begin{itemize}
\item Build a scalable, multilingual AI agent that provides real-time, context-aware repair guidance.
\item Integrate a global technician knowledge-sharing forum to complement automated responses.
\item Test the tool's accuracy, relevance, and usability in both clinical and training environments.
\item Evaluate its adaptability across different medical device types and manufacturers.
\end{itemize}

This study does not include physical repair execution, device manufacturing processes, or regulatory approval for device use, which are currently being investigated. Instead, it aims to establish a digitalized and generalizable support framework that can bridge the technical knowledge gap for biomedical equipment across a variety of healthcare settings, with a focus on medical imaging devices.

\section{Methods}
\subsection{Research Design}
This project follows a phased, mixed-methods approach, combining iterative system development with both quantitative system performance evaluations and qualitative user feedback analysis. It is designed to be descriptive and exploratory, establishing the foundational tools, workflows, and interactions needed to support scalable medical device maintenance via AI, particularly in resource-limited healthcare environments. The project is being carried out in 5 phases, outlined briefly below, with the initial proof-of-concept (Phase 0) reported in detail below. 

\begin{figure}
    \centering
    \includegraphics[width=1.0\linewidth]{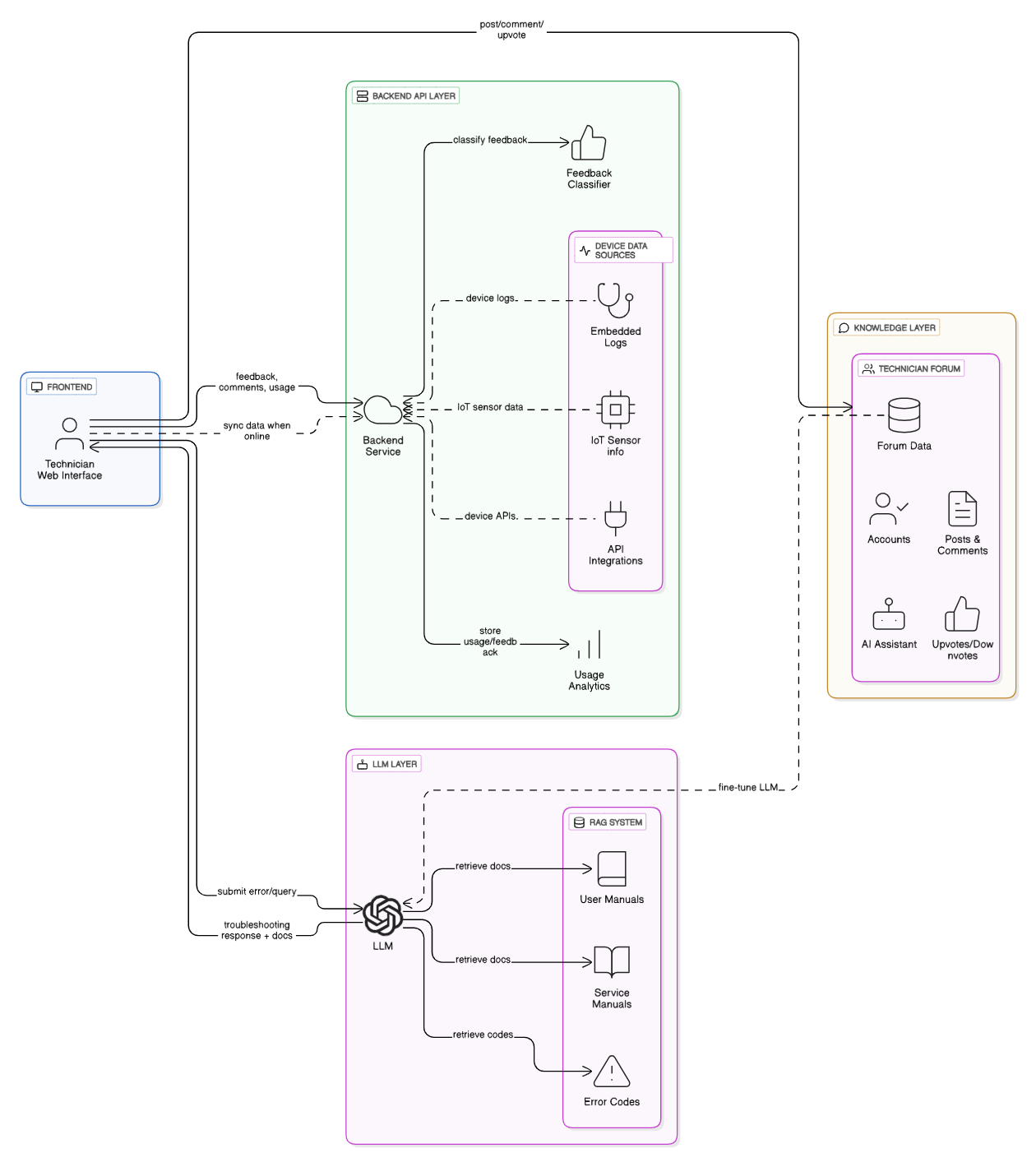}
    \caption{\textbf{INGENZI Tech System Architecture:} 
A high-level overview of the INGENZI Tech platform, showing how technicians interact with a multilingual, offline-capable AI assistant powered by a Retrieval-Augmented Generation (RAG) pipeline, segmented vector stores, and a peer-support forum for real-time troubleshooting and continuous learning.}
    \label{fig:1}
\end{figure}

\subsubsection{Phase 0: Proof of Concept (Completed)}

The initial prototype was developed using the Philips HDI 5000 ultrasound machine as a case study. The assistant was built using a RAG (Retrieval-Augmented Generation) architecture supported by:

\begin{itemize}
    \item \textbf{LLM:} GPT-3.5 Turbo for generation and embedding
    \item \textbf{Vector store:} Facebook AI Similarity Search (FAISS) for document retrieval
    \item \textbf{Frontend/Backend:} React and Flask
    \item \textbf{Core functionalities:}
    Error code lookup,
    Log analysis,
    Self-test simulation,
    Maintenance scheduling via Google Calendar API,
    Multilingual chatbot interface
\end{itemize}

To improve retrieval accuracy and reduce hallucinations, separate vector databases were used for User Manuals, Service Manuals, and Error Codes. The proof-of-concept achieved \textbf{100\%} precision in error code matching and \textbf{80\%} troubleshooting accuracy using internal evaluation methods.

\subsubsection{Phase 1: Forum Integration \& Feedback Loop}

This phase introduces a collaborative technician forum to complement the AI assistant. The forum will allow users to post issues, share solutions, upvote helpful responses, and generate valuable context data.

\textbf{Development Tasks:}

Build or integrate a forum engine
\begin{itemize}
    \item Link forum user accounts with diagnostic tool profiles
    \item Design a schema for tracking usage and technician feedback
    \item Connect forum feedback into a structured pipeline for LLM fine-tuning and retrieval augmentation
\end{itemize}

This human-in-the-loop design ensures continuous model improvement and crowd-sourced domain knowledge.

\subsubsection{Phase 2: API \& IoT Connectivity for Device Integration}

The second development phase focuses on real-time interaction with physical devices, enabling automated log streaming and predictive diagnostics.

\textbf{Planned Tasks:}
\begin{itemize}
    \item Evaluate and implement healthcare communication protocols such as DICOM, HL7, and MQTT.
    \item Define secure API contracts for medical device integration. 
    \item Build a simulation layer for testing log ingestion without live devices. 
    \item Prototype fault-alert dashboards for remote monitoring.
\end{itemize}

This phase extends the system beyond a passive assistant into a proactive maintenance layer.

\subsubsection{Phase 3: Model Optimization \& Continuous Learning}

With data from the tool and forum, the LLM will be fine-tuned for domain-specific reasoning and reliability.

\textbf{Optimization Activities:}
\begin{itemize}
    \item Introduce a feedback classifier to label suggestions as correct/incorrect. 
    \item Use forum responses and real-world interactions for semi-supervised learning. 
    \item Monitor hallucination frequency and optimize vector store indexing. 
    \item Regularly retrain and evaluate the LLM against gold-standard troubleshooting guides. 
\end{itemize}

\subsubsection{Phase 4: Pilot Deployment \& Evaluation}
The system will be deployed and evaluated in clinical environments in LMICs, starting with East Africa.

\textbf{Execution Plan:}
\begin{itemize}
    \item Partner with hospitals and biomedical schools for pilot testing
    \item Train local technicians on usage and provide onboarding materials 
        \item Collect metrics:
        \subitem Time-to-diagnosis
        \subitem Technician satisfaction
        \subitem Device uptime improvement

    \item Use real feedback to further optimize UI, model performance, and workflow integration
\end{itemize}

All user interactions will be logged and anonymized for research analysis. No patient data will be used.

\subsubsection{Phase 5: Multi-Device Expansion}
In this phase, INGENZI Tech will scale to support a broader range of devices (MRI, CT, and X-ray), beginning with one original equipment manufacturer (OEM, i.e., Siemens).

\textbf{Tasks:}
\begin{itemize}
    \item Collect manuals, error codes, and logs for new devices
    \item Expand vector stores or segment them by device class
    \item Add model-switching capabilities in the backend based on device detection
    \item Standardize input/output formats for scalable ingestion
\end{itemize}

This positions the system as a brand-agnostic, AI-powered maintenance layer that can operate across multiple health systems and equipment ecosystems.

\begin{figure}
    \centering
    \includegraphics[width=0.9\linewidth]{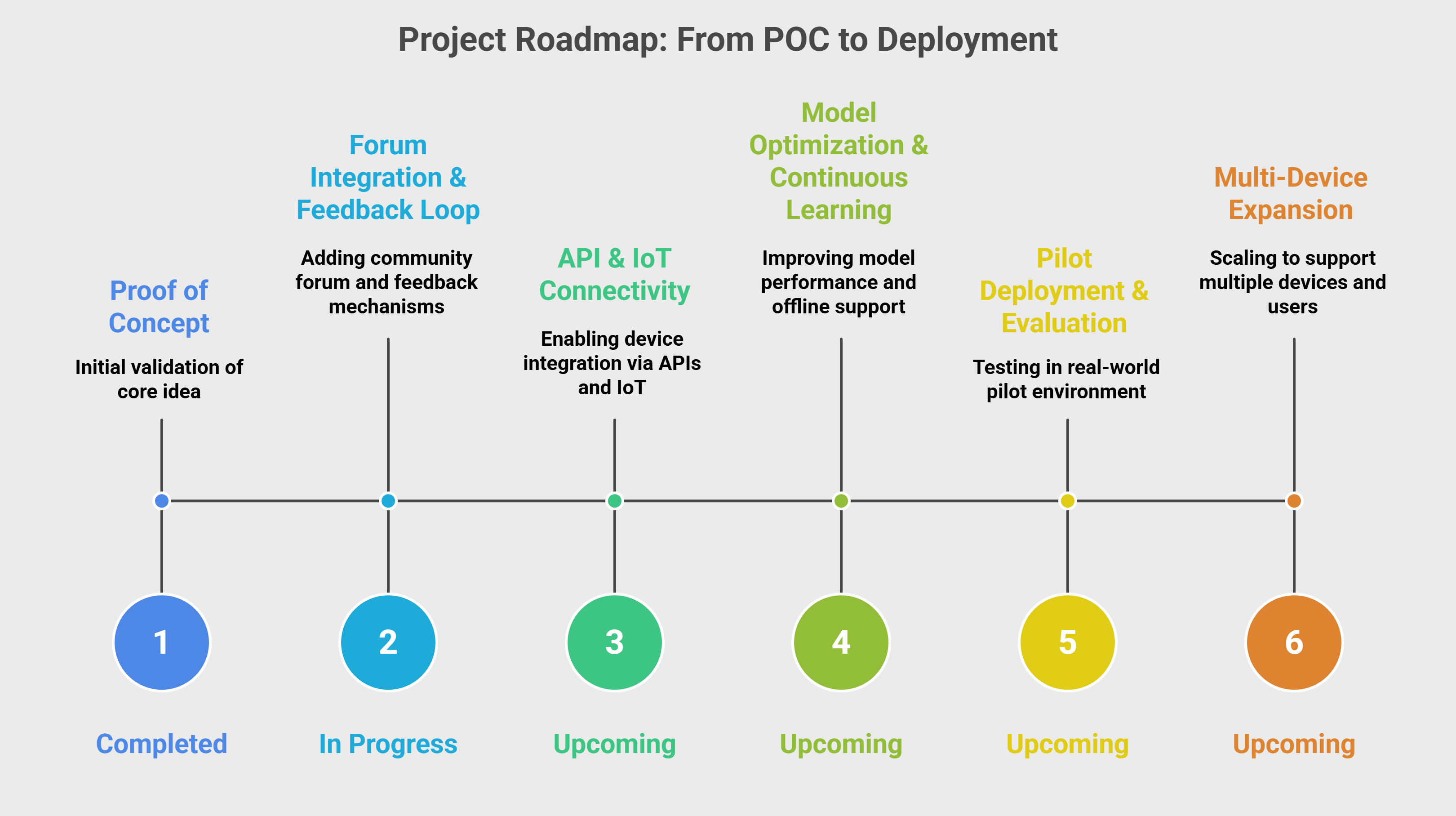}
    \caption{High-level roadmap of the INGENZI Tech project phases, illustrating the evolution from proof-of-concept to scalable multi-device support across LMIC healthcare environments.}
    \label{fig:2}
\end{figure}

\subsection{Phase 0 Dataset}
The dataset used in Phase 0 was built from 15 technical documents related to the Philips HDI 5000 ultrasound system, including user manuals, service manuals, and detailed error code catalogs. These documents were parsed and chunked using a semantic-aware splitting strategy to preserve contextual meaning. Each chunk was then embedded using OpenAI's GPT-3.5 Turbo embedding model and stored in Facebook AI Similarity Search (FAISS),  a high-performance vector database.
FAISS was selected due to its ability to handle large-scale similarity search efficiently, offering fast retrieval speeds with minimal memory overhead. Given the need for real-time, lightweight deployment in resource-constrained settings, FAISS provided the optimal trade-off between scalability, speed, and customization. To further reduce hallucinations and improve specificity, the vector database was segmented into three discrete stores: (1) user manuals, (2) service manuals, and (3) error codes.
This structured storage approach allowed for domain-specific querying and significantly improved the system's contextual accuracy during troubleshooting tasks.

\subsection{LLM Model}

The Phase 0 assistant was built using the GPT-3.5 Turbo model and operated within a Retrieval-Augmented Generation (RAG) framework. Technician queries were first interpreted through prompt engineering and sent to the FAISS-powered vector stores to retrieve the top-k most relevant document chunks. These chunks were then appended as context to the LLM prompt, enabling accurate and explainable response generation.
The system supported various custom tools, including: \textit{Error code lookup} \textit{Log parsing}, \textit{Maintenance schedule generation}, and \textit{Simulated self-testing}

Each tool selection was driven by prompt conditioning and intent classification strategies. The backend was deployed using a lightweight Flask server, interfacing with a React-based frontend, ensuring accessibility on low-power machines.
By combining a modular tool pipeline with FAISS-enhanced RAG retrieval, the system maintained sub-10-second average latency, making it suitable for near real-time use in clinical environments.

\section{Preliminary Results of Phase 0}

To evaluate the effectiveness of the INGENZI Tech prototype, we conducted two core assessments focused on the platform's accuracy in retrieving device-specific information using its RAG-based architecture.

\subsection{Error Code Interpretation Accuracy}

A total of 90 error codes, along with their corresponding descriptions, were extracted from the Philips HDI 5000 ultrasound service manuals. These were used as input queries to the system to test its ability to retrieve accurate contextual matches from the embedded vector stores. The system successfully retrieved the correct description for \textbf{100\% of the error codes}, confirming high alignment between query semantics and stored content. This demonstrated the model's robustness in precise retrieval of structured, short-form diagnostic information.

\subsection{Instructional Query Evaluation}

To assess the system's performance on broader instructional support, we curated a set of 30 natural language queries derived from the device's user and service manuals. These questions spanned operational instructions, safety guidelines, and device handling procedures. Out of the 30 questions, the system returned accurate and complete responses for 24 of them, yielding an \textbf{80\% success rate}. The remaining six responses were either partially accurate or missing critical details, indicating areas for improvement in document chunking, relevance scoring, or retrieval granularity.

\subsection{Implications}

These preliminary results validate the feasibility of leveraging a segmented RAG framework for technical document retrieval in low-resource settings. The perfect performance on structured error code queries suggests high utility in fault identification workflows, while the 80\% accuracy on unstructured instructional queries highlights potential for effective user guidance, with room for refinement in LLM prompt tuning and vector search thresholds.

\begin{figure}
    \centering
    \includegraphics[width=0.9\linewidth]{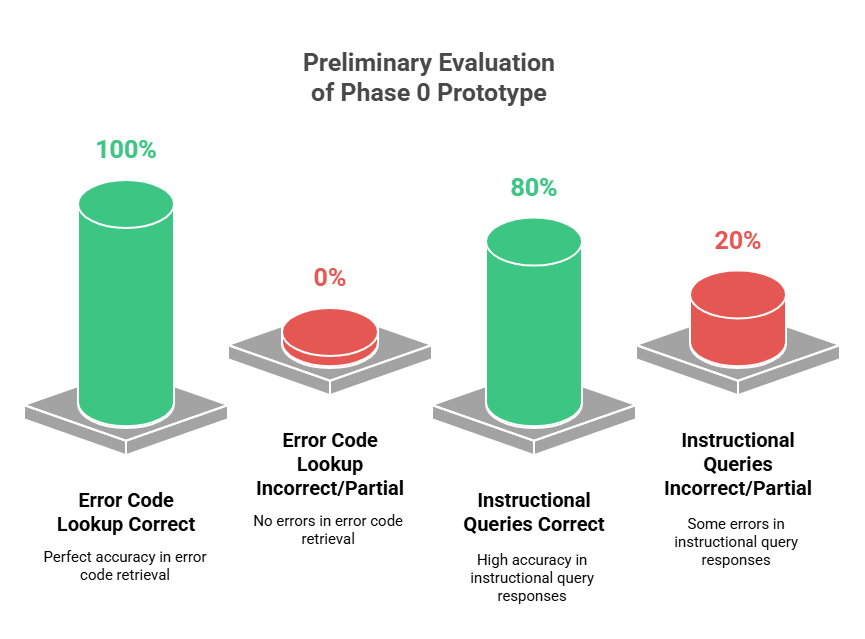}
    \caption{Performance comparison between structured (error codes) and unstructured (instructional) query handling by the INGENZI Tech prototype.}
    \label{fig:3}
\end{figure}

\section{Discussion}

This study is expected to yield a robust, AI-powered support platform tailored to the needs of biomedical technicians in low-resource settings. The anticipated outcomes include both tangible software components and insights that can inform future research and development in the domain of intelligent medical device maintenance.
First, the research aims to validate the effectiveness of a multilingual diagnostic assistant powered by large language models (LLMs). This assistant will interpret error codes, retrieve context-specific documentation, and guide users through actionable repair workflows. Evaluation will rely not only on internal metrics but also on established LLM benchmarks adapted to the task of technical reasoning and document retrieval. Performance targets include high exact match (EM), precision, and recall in context matching, as well as positive assessments from real-world users.
To enhance accuracy and reduce hallucinations, the system will incorporate segmented vector stores, specifically dedicated to user manuals, service manuals, and error codes. This structural refinement is expected to result in more contextually relevant responses and better model alignment with technician expectations.
In addition to the AI assistant, the project will deliver a fully integrated peer-to-peer discussion forum. This forum will allow technicians to post unresolved issues, upvote helpful responses, and share field-tested solutions. Feedback gathered through this platform will contribute to a semi-supervised learning loop for continuous model refinement.
A key outcome of this study is ensuring accessibility. The system will include offline support and a multilingual interface to accommodate technicians working in bandwidth-constrained or non-English-speaking environments, typical of many LMICs.
Looking ahead, the solution will be structured for scalability across multiple device types. While the proof of concept focused on the Philips HDI 5000 ultrasound machine, the system architecture will support transition to more advanced diagnostic tools such as Siemens MRI, CT, and X-ray systems. This includes the addition of device-detection logic and ingestion format standardization to support multi-device workflows.
Finally, the research will provide user insights drawn from technician interaction logs, surveys, and qualitative feedback. These insights will inform improvements in AI-guided repair recommendations, interface usability, and system reliability in real-world clinical contexts.

\textbf{Final Deliverables:}
\begin{itemize}
    \item A fully functional, web-based diagnostic tool powered by an LLM and Retrieval-Augmented Generation (RAG)
    \item A segmented vectorized knowledge base of user manuals, service guides, and error codes.
    \item A technician discussion forum with structured feedback tracking.
    \item Offline-capable and multilingual user interface.
    \item A comprehensive evaluation report, including benchmark results and technician feedback. 
    \item Technical documentation, onboarding materials, and a deployment-ready software package for pilot testing
\end{itemize}

\section{Significance of the Study}

This study addresses a pressing and persistent gap in global health systems, ensuring that life-saving medical devices remain operational in low-resource environments. In LMICs, prolonged equipment downtime due to a lack of technical expertise and support results in avoidable diagnostic delays, poor patient outcomes, and significant resource waste. By offering a scalable, AI-powered support platform tailored to biomedical technicians, this research directly contributes to the continuity and quality of healthcare services where they are most vulnerable.
The proposed solution aligns with global health and technology goals, including WHO's call for improved health technology management and the increasing emphasis on sustainable, equitable digital health infrastructure. Unlike many existing tools limited to specific manufacturers or high-connectivity settings, this platform provides multilingual, offline-capable assistance and incorporates peer knowledge exchange to support long-term learning.
By combining cutting-edge natural language processing, practical engineering workflows, and feedback-driven model improvement, the study also contributes to the growing field of human-centered AI in healthcare. Its outcomes can inform the development of future AI tools for technical domains that suffer from knowledge bottlenecks, especially in under-resourced contexts.
Ultimately, this research not only advances technical innovation but has the potential to transform biomedical maintenance from a reactive to a proactive discipline, reducing equipment failure, empowering local technicians, and improving care delivery across LMIC health systems.

\section{Conclusion}
This research proposes a practical and scalable solution to a longstanding challenge in global healthcare, ensuring the continuous functionality of medical devices in low-resource settings. By combining large language models, structured knowledge retrieval, and collaborative technician input, the platform aims to shift maintenance practices from reactive to proactive. The integration of multilingual, offline support ensures accessibility, while modular design allows expansion to various device types and manufacturers. Through phased development and real-world validation, this study not only contributes a novel AI tool but also advances the broader mission of equitable, reliable healthcare delivery in LMICs and beyond.


\newpage

\bibliographystyle{splncs04}
\bibliography{reference}

\end{document}